\documentclass{article} 
\usepackage{iclr2016_conference,times}
\usepackage{subfigure}
\usepackage{multirow}
\usepackage{graphicx}
\usepackage{color}

\iclrfinalcopy

\title{Better Computer Go Player with Neural Network and Long-term Prediction}
\author{Yuandong Tian\\
Facebook AI Research \\
Menlo Park, CA 94025 \\
\texttt{yuandong@fb.com}
\And 
Yan Zhu \\
Rutgers University\\
Facebook AI Research \\
\texttt{yz328@cs.rutgers.edu}
}

\def\newcite#1{[\cite{#1}]}
\def\url#1{#1}

\definecolor{MyDarkBlue}{rgb}{0,0.1,1}

\definecolor{YanColor}{rgb}{1,0,0}

\date{November 2015}

\begin{document}
\maketitle

\begin{abstract}
    Competing with top human players in the ancient game of Go has been a long-term goal of artificial intelligence. Go's high branching factor makes traditional search techniques ineffective, even on leading-edge hardware, and Go's evaluation function could change drastically with one stone change. Recent works~\newcite{deepmind,edinburgh} show that search is not strictly necessary for machine Go players. A pure pattern-matching approach, based on a Deep Convolutional Neural Network (DCNN) that predicts the next move, can perform as well as Monte Carlo Tree Search (MCTS)-based open source Go engines such as Pachi~\newcite{pachi} if its search budget is limited. We extend this idea in our bot named \emph{darkforest}, which relies on a DCNN designed for long-term predictions. \emph{Darkforest} substantially improves the win rate for pattern-matching approaches against MCTS-based approaches, even with looser search budgets. Against human players, the newest versions, \emph{darkfores2}, achieve a stable \emph{3d level} on KGS Go Server as a ranked bot, a substantial improvement upon the estimated 4k-5k ranks for DCNN reported in~\cite{edinburgh} based on games against other machine players. Adding MCTS to \emph{darkfores2} creates a much stronger player named \emph{darkfmcts3}: with 5000 rollouts, it beats Pachi with 10k rollouts in all 250 games; with 75k rollouts it achieves a stable \emph{5d level} in KGS server, on par with state-of-the-art Go AIs (e.g., Zen, DolBaram, CrazyStone) except for AlphaGo~\newcite{nature2016deepmind}; with 110k rollouts, it won the 3rd place in January KGS Go Tournament.
\end{abstract}

\section{Introduction}
\begin{figure}
\centering
\includegraphics[width=0.95\textwidth]{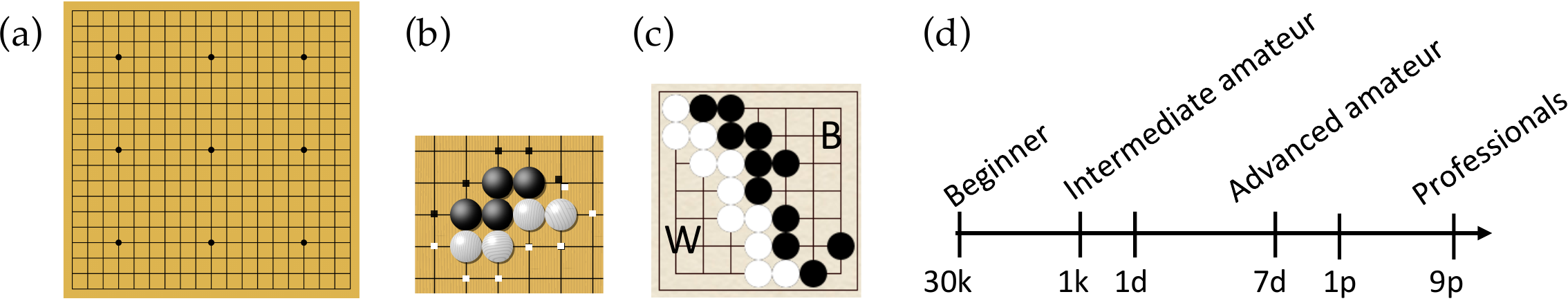}
\caption{A simple illustrations on Go rules and rating system. Images are from Internet.}
\label{fig:go-intro}
\end{figure}

For a long time, computer Go is considered to be a grand challenge in artificial intelligence. Fig.~\ref{fig:go-intro} shows a simple illustration of the game of Go. Two players, black and white, place stones at intersections in turn on a 19x19 board (Fig.~\ref{fig:go-intro}(a)). Black plays first on an empty board. A 4-connected component of the same color is called a \emph{group}. The \emph{liberties} of a group is the number of its neighboring empty intersections (Fig.~\ref{fig:go-intro}(b)). A group is \emph{captured} if its liberties is zero. The goal of the game is to control more territory than the opponent (Fig.~\ref{fig:go-intro}(c)). Fig.~\ref{fig:go-intro}(d)) shows the Go rating system, ranging from \emph{k}yu level (beginner to decent amateur, 30k-1k) to \emph{d}an level (advanced amateur, 1d-7d) and to professional levels (1p-9p)~\newcite{silver2009reinforcement}.

Go is difficult due to its high branching factors (typically on the order of hundred on a 19x19 board) and subtle board situations that are sensitive to small changes (adding/removing one stone could alter the life/death situation of a large group of stone and thus completely changes the final score). A combination of the two implies that the only solution is to use massive search that requires a prohibitive amount of resources, which is not attainable with cutting-edge hardware.

Fortunately, recent works~\newcite{deepmind, edinburgh} in Computer Go have shown that the Go board situation could be deciphered with Deep Convolutional Neural Network (DCNN). They can predict the next move that a human would play $55.2\%$ of the time. However, whether this accuracy leads to a strong Go AI is not yet well understood. It is possible that DCNN correctly predicts most regular plays by looking at the correlation of local patterns, but still fails to predict the critical one or two moves and loses the game. Indeed, a DCNN-based player is still behind compared to traditional open-source engines based on Monte-Carlo Tree Search (MCTS)~\newcite{browne2012survey,kocsis2006bandit}, let alone commercial ones.

In this paper, we show that DCNN-based move predictions indeed give a strong Go AI, if properly trained. In particular, we carefully design the training process and choose to predict next $k$ moves rather than the immediate next move to enrich the gradient signal. Despite our prediction giving a mere 2\% boost for accuracy of move predictions, the win rate against open-source engines (e.g., Pachi and Fuego) in heavy search scenarios (e.g., 100k rollouts) is \emph{more than 6 times higher} (Pachi: $11.0\%$ vs $72.6\%$, Fuego: $12.5\%$ vs $89.7\%$) than current state-of-the-art DCNN-based player~\newcite{deepmind}. In addition, our search-less bot \emph{darkfores2} played ranked game on KGS Go Server and achieves stable 3d level, much better than the neural network based AI proposed by~\cite{edinburgh} that holds 4-5 kyu estimated from games against other MCTS-based Go engines.

Our bots also share the common weakness of DCNN-based methods in local tactics. Combining DCNN with MCTS, our hybrid bot \emph{darkfmcts3} addresses such issues. With 5000 rollouts, it beats Pachi with 10k rollouts in all 250 games ($100\%$ win rate); with 75k rollouts it achieves a stable \emph{5d level} in KGS Go server, on par with state-of-the-art Go AIs (e.g., Zen, DolBaram, CrazyStone); with 110k rollouts, it won the 3rd place in January KGS Go Tournament. Recently, DeepMind's AlphaGo~\newcite{nature2016deepmind} defeated European Go Champion Fan Hui with 5-0, showing the strong power of DCNN-based bot.

\section{Method}
Using Neural Network as a function approximator and pattern matcher to predict the next move of Go is a long-standing idea~\newcite{sutskever2008mimicking,richards1998evolving,schraudolph1994temporal,enzenberger1996integration}. Recent progress~\newcite{deepmind,edinburgh} uses \emph{Deep} Convolutional Neural Network (DCNN) for move prediction, and shows substantial improvement over shallow networks or linear function approximators based on manually designed features or simple patterns extracted from previous games~\newcite{silver2009reinforcement}.

In this paper, we train a DCNN that predicts the next $k$ moves given the current board situation as an input. We treat the $19\times 19$ board as a $19\times 19$ image with multiple channels. Each channel encodes a different aspect of board information, e.g., liberties (Fig.~\ref{fig:go-intro}(b)). Compared to previous works, we use a more compact feature set and predict long-term moves, and show that they lead to a substantial performance boost in terms of win rate against open source engines. 

\subsection{Feature Channels}
\begin{figure}
\centering
\vspace{-0.1in}
\includegraphics[width=0.95\textwidth]{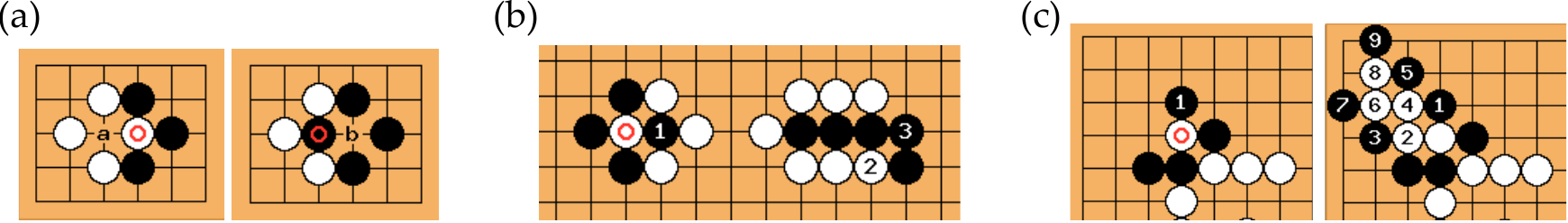}
\caption{Some special situations in Go. \textbf{(a)} \emph{Ko}. After black captures white stone by playing at \emph{a}, white is prohibited to capture back immediately by playing at \emph{b} to prevent repetition of game state. \textbf{(b)} \emph{Ko fight}. Black captures white at 1, white cannot capture back. Instead, white can plays at 2, threatening the three black stones (called \emph{Ko threat}). If black plays at 3 to connect, white can then win back the Ko. \textbf{(c)} \emph{Ladder}. Black plays at 1, threatening to capture the white stone at circle. White escapes but eventually gets captured at the border. Each time after black plays, white's liberties shrink from 2 to 1. Images from Sensei's Library (\url{http://senseis.xmp.net/}).}
\label{fig:special-situation}
\end{figure}

\begin{table}[h]
\small
\centering
\begin{tabular}{c||c|c|c|c}
                 & Name & Type & Description & \#planes \\
\hline\hline
\multirow{5}{*}{standard} &  our/opponent liberties & binary & \textbf{true} if the group has 1, 2 and $\ge 3$ liberties & 6 \\
& Ko (See Fig.~\ref{fig:special-situation}(a)) & binary & \textbf{true} if it is a Ko location (illegal move) \hfill & 1 \\
& our/opponent stones/empty & binary & - & 3 \\
& our/opponent history & real & how long our/opponent stone is placed & 2 \\
& opponent rank & binary & All \textbf{true} if opponent is at that rank & 9 \\
\hline\hline
\multirow{3}{*}{extended} & border & binary & \textbf{true} if at border & 1 \\
& position mask & real & $\exp(-.5*distance^2)$ to the board center  & 1 \\
& our/opponent territory & binary  & \textbf{true} if the location is closer to us/opponent. & 2 \\ 
\end{tabular}
\caption{Features extracted from the current board situation as the input of the network. Note that extended feature set also includes standard set. As a result, standard set has 21 channels while extended one has 25 channels.}
\label{tab:feature-list}
\end{table}

Table~\ref{tab:feature-list} shows the features extracted from the current board situation. Each feature is a binary $19\times 19$ map except for history information and position mask, which are real numbers in $[0, 1]$. History is encoded as $\exp(-t*0.1)$, where $t$ is how long the stone has been placed. The exponential temporal decay is meant to enable the network to focus on the recent battle. Position mark is defined as $\exp(-\frac{1}{2}l^2)$, where $l^2$ is the squared L2 distance to the board center. It is used to encode the relative position of each intersection.

There are two differences between our features and those in~\cite{deepmind}. First, we use relative coding (our/opponent) for almost all the features. In contrast, the features in~\cite{deepmind} are largely player-agnostic. Second, our feature set is simpler and compact (25 vs. 36 input planes), in particular, free from one step forward simulation. In comparison,~\cite{deepmind} uses such features like liberties after the move, captures after the move, etc. 

We use a similar way to encode rank in 9 planes as in~\cite{deepmind}. That is, all kyu-players have all nine planes zero, 1d players has their first plane all-1, 2d players have their second plane all-1, etc. For 9d and professional players, all the planes are filled with $1$. 

\begin{figure}
\centering
\vspace{-0.1in}
\includegraphics[width=0.95\textwidth]{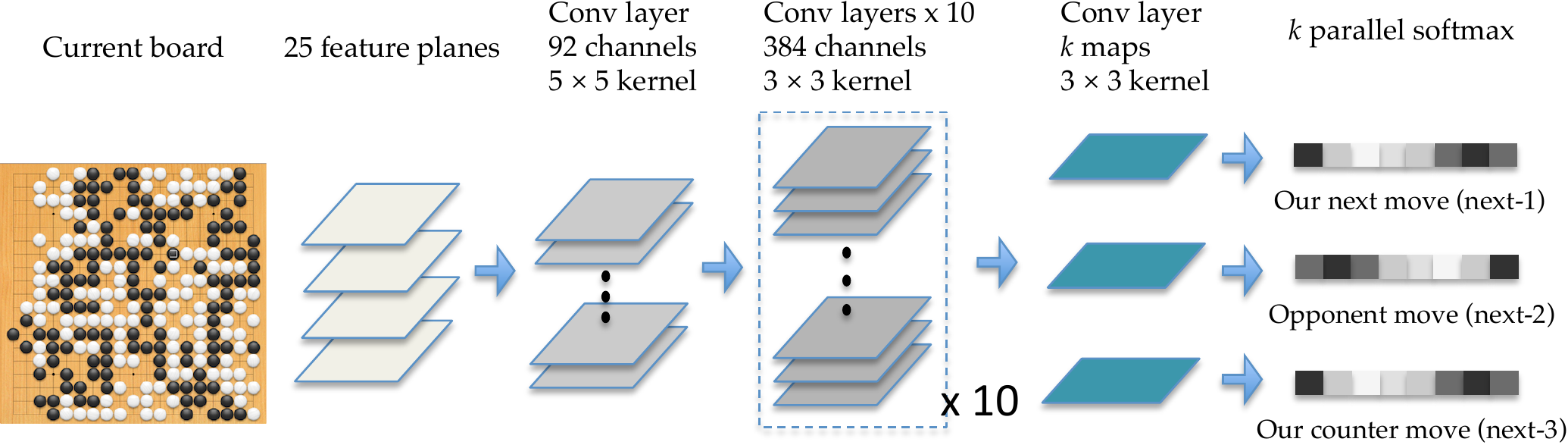}
\caption{Our network structure ($d=12$, $w=384$). The input is the current board situation (with history information), the output is to predict next $k$ moves.}
\label{fig:network-arch}
\end{figure}

\subsection{Network Architecture}
Fig.~\ref{fig:network-arch} shows the architecture of the network for our best model. We use a 12-layered ($d=12$) full convolutional network. Each convolution layer is followed by a ReLU nonlinearity. Except for the first layer, all layers use the same width $w=384$. No weight sharing is used. We do not use pooling since they negatively affect the performance. Instead of using two softmax outputs~\newcite{deepmind} to predict black and white moves, we only use one softmax layer to predict the next move, reducing the number of parameters.

\subsection{Long term planning}
Predicting only the immediate next move limits the information received by the lower layers. Instead, we predict next $k$ moves (self and opponent, alternatively) from the current board situation. Each move is a separate softmax output. The motivation is two-fold. First, we want our network to focus on a strategic plan rather than the immediate next move. Second, with multiple softmax outputs, we expect to have more supervisions to train the network. Table~\ref{tab:gradient-ratio} computes the ratio of average gradient L2 norm (over the first 9 epochs, first 1000 mini-batches removed) between 1-step and 3-step predictions at each convolutional layer. As expected, the gradient magnitudes of the top layers (layers closer to softmax) are higher in 3-step prediction. However, the gradient magnitudes of the lower layers are approximately the same, showing that the lower gradients are canceled out in 3-step prediction, presumably leaving only the most important gradient for training. Empirically, DCNN trained with 3 steps gives high win rate than that with 1 step. 

\begin{table}[h]
\centering
\begin{tabular}{c||c|c|c|c|c|c}
                         layer &  conv1 & conv3 & conv5 & conv7 & conv9 & conv11\\
\hline\hline
gradient norm ratio     & 1.18  &  1.20    &  1.37  &  1.46  &  1.98  &  2.41       \\

\end{tabular}
\caption{Comparison in gradient L2 norm between 1-step prediction and 3-step prediction network.}
\label{tab:gradient-ratio}
\end{table}

\subsection{Training}
When training, we use 16 CPU threads to prepare the minibatch, each simulating 300 random selected games from the dataset. In each minibatch, for each thread, randomly select one game out of 300, simulate one step according to the game record, and extract features and next $k$ moves as the input/output pair in the batch. If the game has ended (or fewer than $k$ moves are left), we randomly pick one (with replacement) from the training set and continue. The batch size is $256$. We use data augmentation with rotation at 90-degree intervals and horizontal/vertical flipping. For each board situation, data augmentation could generate up to 8 different situations.

Before training, we randomly initialize games into different stages. This ensures that each batch contains situations corresponding to different stages of games. Without this, the network will quickly overfit and get trapped into poor local minima. 

Because of our training style, it is not clear when the training set has been thoroughly processed once. Therefore, we just define an epoch as 10,000 mini-batches. Unlike~\cite{deepmind} that uses asynchronous stochastic gradient descent, we just use vanilla SGD on 4 NVidia K40m GPUs in a single machine to train the entire network (for some models we use 3 GPUs with 255 as the batch size). Each epoch lasts about 5 to 6 hours. The learning rate is initially 0.05 and then divided by 5 when convergence stalls. Typically, the model starts to converge within one epoch and shows good performance after 50-60 epochs (around two weeks). 

Other than simplest DCNN model, we also tried training with ResNet [\cite{he2015deep}] which recently gives state-of-the-art performance in image classification. We also tried using additional targets, such as predicting the endgame territories given the current board status. Both gives faster convergence. Recurrent Neural Network is also tried but gives worse performance. 

\subsection{Monte Carlo Tree Search}
\label{sec:mcts-intro}
\begin{figure}
\centering
\vspace{-0.1in}
\includegraphics[width=0.95\textwidth]{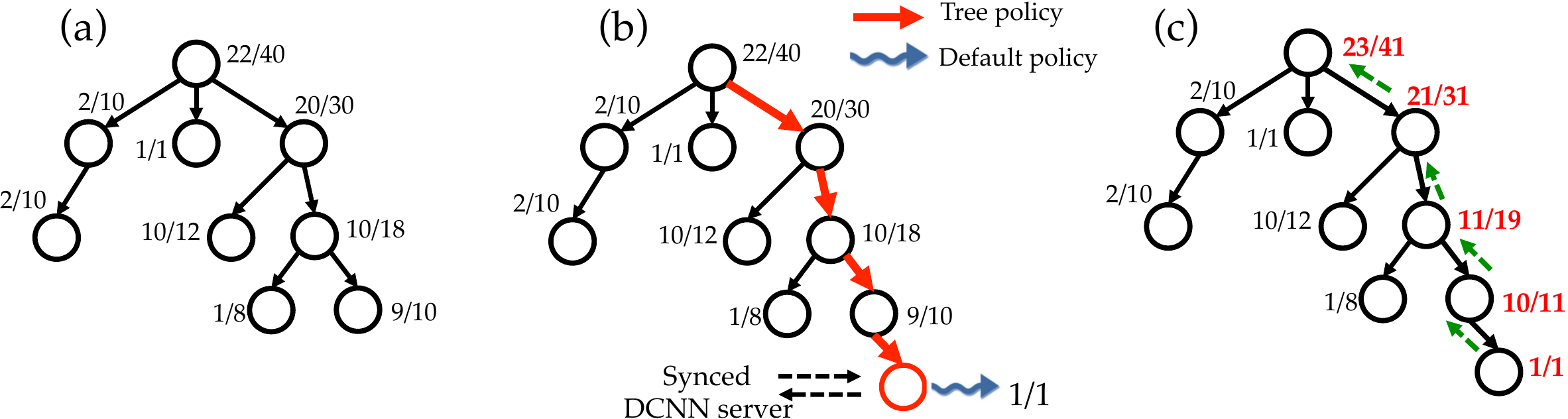}
\caption{A brief illustration of MCTS with DCNN. \textbf{(a)} A game tree. For each node, the statistics $m/n$ indicates that from the node, $n$ games are emulated, out of which $m$ are won by black. Root represents the current game state. \textbf{(b)} A new rollout starting from the root. It picks a move from the current state using \emph{tree policy} and advances to the next game state, until it picks the a new move and expand a new leaf. From the leaf, we run \emph{default policy} until the game ends (black wins in the illustration). At the same time, the leaf status is sent to a DCNN server for evaluation. For synchronized implementation, this new node is available for tree policy after the evaluation is returned. \textbf{(c)} The statistics along the trajectory of the tree policy is updated accordingly.}
\vspace{-0.1in}
\label{fig:mcts-intro}
\end{figure}

From the experiments, we clearly show that DCNN is tactically weak due to the lack of search. Search is a way to explore the solution space conditioned on the current board situation, and build a non-parametric local model for the game. The local model is more flexible than the global model learned from massive training data and more adapted to the current situation. The state-of-the-art approach in computer Go is Monte-Carlo Tree Search (MCTS). Fig.~\ref{fig:mcts-intro} shows its basic principle.

Combining DCNN with MCTS requires nontrivial engineering efforts because each rollout of MCTS is way much faster than DCNN evaluation. Therefore, these two must run in parallel with frequent communications. Our basic implementation of MCTS gives 16k rollouts per second (for 16 threads on a machine with Intel Xeon CPU E5-2680 v2 at 2.80GHz) while it typically takes 0.2s for DCNN to give board evaluations of a batch size of 128 with 4 GPUs. 

There are two ways to address this problem. In \emph{asynchronized implementation} used in~\cite{deepmind}, MCTS sends the newly expanded node to DCNN but is not blocked by DCNN evaluation. MCTS will use its own tree policy until DCNN evaluation arrives and updates moves. This gives high rollout rate, but there is a time lag for the DCNN evaluation to take effect, and it is not clear how many board situations have been evaluated for a given number of MCTS rollouts. In \emph{synchronized implementation}, MCTS will wait until DCNN evaluates the board situation of a leaf node, and then expands the leaf. Default policy may run before/after DCNN evaluation. This is much slower but guarantees that each node is expanded using DCNN's high-quality suggestions.

In our experiments, we evaluate the synchronized case, which achieves $84.8\%$ win rate against its raw DCNN player with only 1000 rollouts. Note that our implementation is not directly comparable to the asynchronized version in~\cite{deepmind}, achieving $86.7\%$ with 100k rollouts. In their recent AlphaGo system~\newcite{nature2016deepmind}, a fast CPU-based tree policy is used before DCNN evaluation arrives. Combined with their default policy that gives a move suggestion in 2$\mu$ s with $24.2\%$ accuracyon a Tygem dataset, the CPU-only system already achieves 3d level. We also try implementing it using similar local patterns and achieves slight higher accuracy ($25.1\%$ evaluated on GoGoD) with 4-5 $\mu$ s per move. However, when combined with our system, its performance is not as good as Pachi's rule-based default policy, showing that top-1 accuracy may not be a sensitive metric to use. Indeed, some moves are much more critical than others. 

\section{Experiments}
\subsection{Setup}
We use the public KGS dataset ($\sim$170k games), which is used in~\cite{deepmind}. We use all games before 2012 as the training set and 2013-2015 games as the test set. This leads to 144,748 games for training and 26,814 games for testing. We also use GoGoD dataset\footnote{We used GoGoD 2015 summer version, purchased from \url{http://www.gogod.co.uk}. We skip ancient games and only use game records after 1800 AD.} ($\sim$80k games), which is also used in~\cite{edinburgh}. 75,172 games are used for training and 2,592 for testing.

For evaluation, our model competes with GnuGo, Pachi~\newcite{pachi} and Fuego~\newcite{fuego}. We use GnuGo 3.8 level 10, Pachi 11.99 (Genjo-devel) with the pattern files, and Fuego 1.1 throughout our experiments.

\subsection{Move prediction}
Table~\ref{tab:move-prediction} shows the performance comparison for move prediction. For models that predict the next $k$ moves, we only evaluate their prediction accuracy for the immediate next move, i.e., the first move the model predicts.

\begin{figure*}[t!]
    \centering
    \begin{subfigure}
        \centering
        \includegraphics[width=0.45\textwidth]{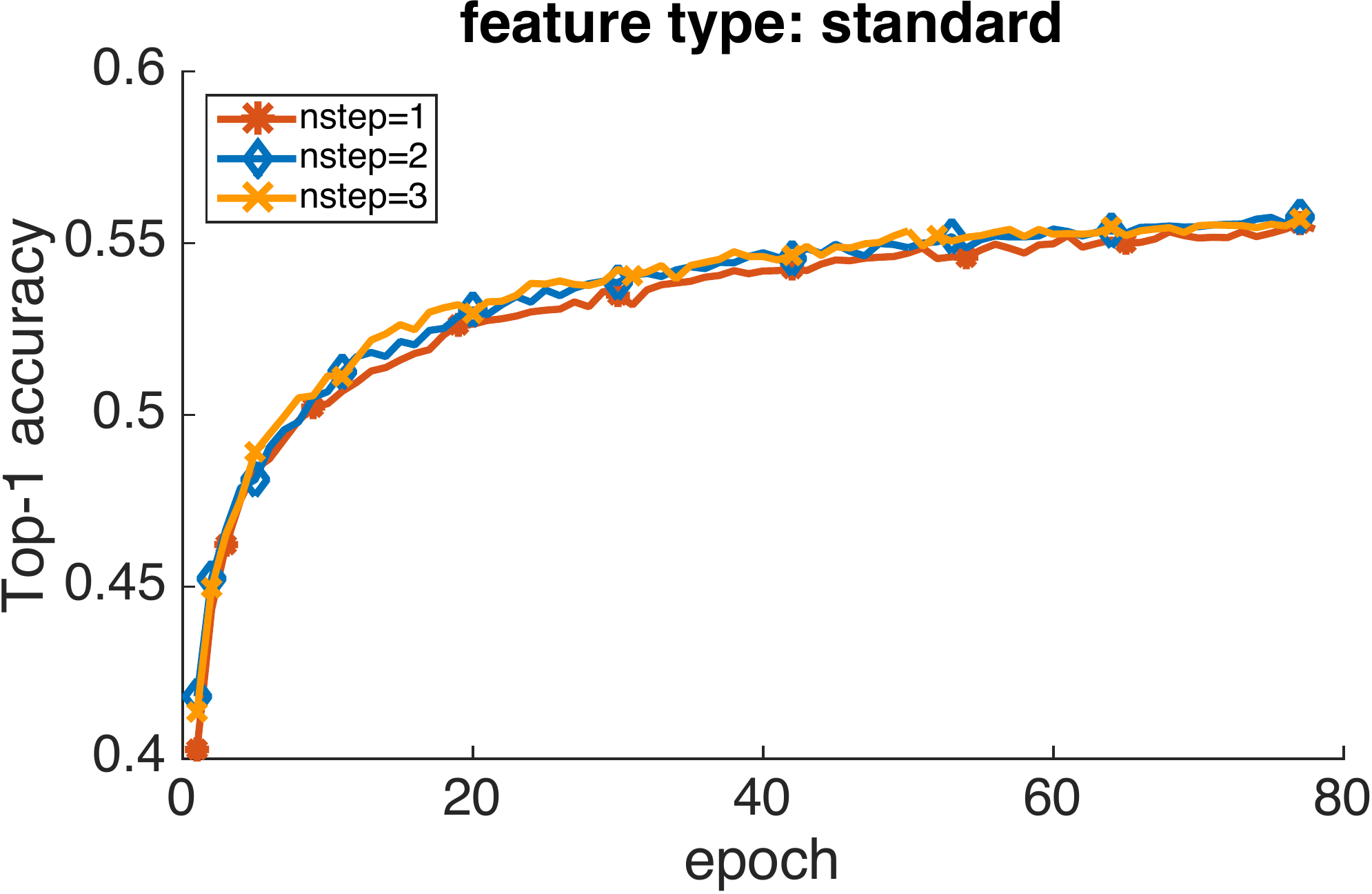}
    \end{subfigure}%
    ~ 
    \begin{subfigure}
        \centering
        \includegraphics[width=0.45\textwidth]{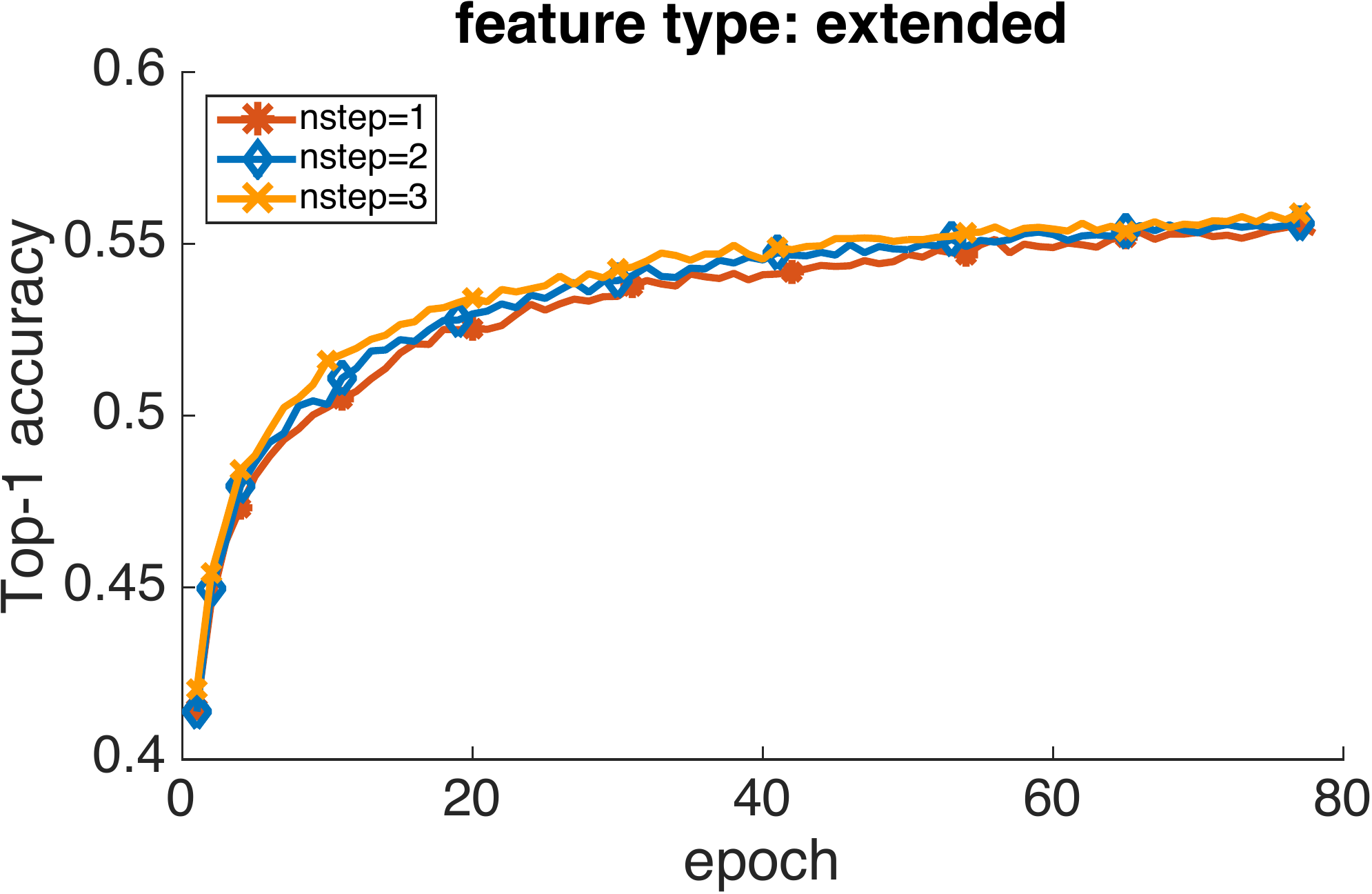}
    \end{subfigure}
    \caption{Top-1 accuracy of the immediate move prediction with $k=1$, $2$ and $3$ next move predictions.}
    \label{fig:testerr}
\end{figure*}

\begin{table}[h]
\centering
\begin{tabular}{c||c|c|c|c}
\cite{deepmind} & d=12,w=384 & d=12,w=512 & d=16,w=512 & d=17,w=512  \\
\hline\hline
55.2 & 57.1 & 57.3 & 56.6 & 56.4 \\
\end{tabular}
\caption{Comparison of Top-1 accuracies for immediate move predictions using standard features. $d$ is the model depth while $w$ is the number of filters at convolutional layers (except the first layer).}
\label{tab:move-prediction}
\end{table}
With our training framework, we are able to achieve slightly higher Top-1 prediction accuracy of immediate next move (after hundreds of epochs) compared to~\cite{deepmind}. Note that using standard or extended features seem to have marginal gains (Fig.~\ref{fig:testerr}). For the remaining experiments, we thus use $d=12$ and $w=384$, as shown in Fig.~\ref{fig:network-arch}.

\subsection{Win rate}
\begin{figure*}[t!]
    \centering
    \begin{subfigure}
        \centering
        \includegraphics[width=0.45\textwidth]{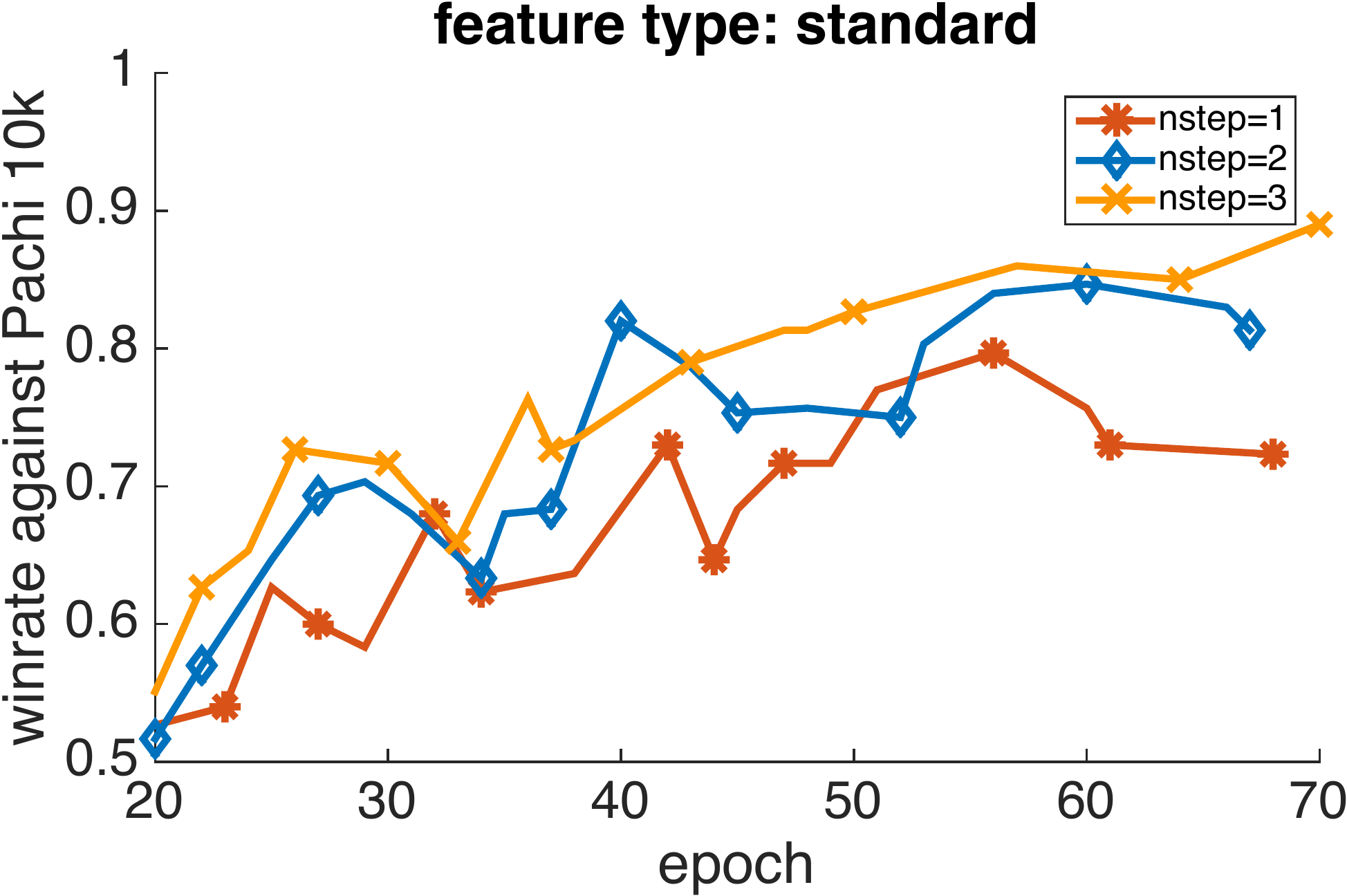}
    \end{subfigure}%
    ~ 
    \begin{subfigure}
        \centering
        \includegraphics[width=0.45\textwidth]{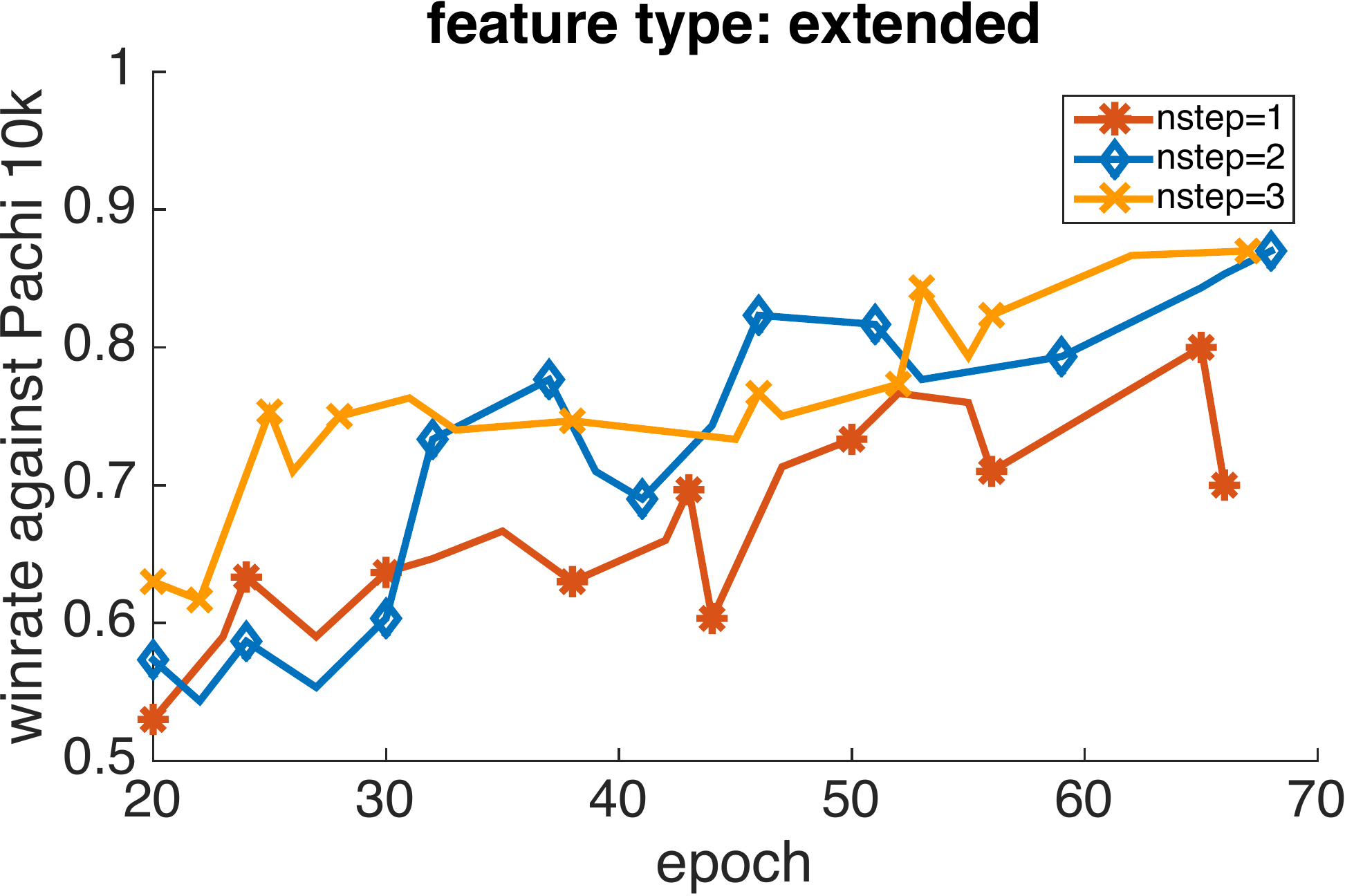}
    \end{subfigure}
    \caption{Evolution of win rate versus Pachi 10k. Each win rate is computed from 300 games.}
    \label{fig:winrate}
\end{figure*}

Although the improvement in move prediction accuracy is small, the improvement in play strength, in terms of win rate, is much larger. Fig.~\ref{fig:winrate} shows the improvement of win rate over time. Our DCNN trained with 2 or 3 steps is about $10\%-15\%$ (in absolute difference) better than DCNN trained with 1 step. More steps show diminishing returns. On the other hand, the win rate of the standard feature set is comparable to the extended one. Table~\ref{tab:wr-best} shows that win rate of our approach is substantially higher than that of previous works. We also train a smaller model with $w=144$ whose number of parameters are comparable to~\cite{deepmind}. Our smaller model achieves $43.3\%$ in 300 games against Pachi 10k when Pachi's pondering is on (keep searching when the opponent plays), and $55.7\%$ when it is off. In contrast,~\cite{deepmind} reports $47.4\%$ and does not mention pondering status.

\textbf{Darkforest AI Bots.} We build three bots from the trained models. Our first bot \emph{darkforest} is trained using standard features, 1 step prediction on KGS dataset. The second bot \emph{darkfores1} is trained using extended features, 3 step prediction on GoGoD dataset. Both bots are trained with constant learning rate $0.05$. Based on \emph{darkfores1}, we fine-tuned the learning rate to create an even stronger DCNN player, \emph{darkfores2}. Note that fine-tuning KGS model achieves comparable strength. Table~\ref{tab:wr-best} shows their strengths against open source engines. It seems that despite the fact that GoGoD is smaller, our model can be trained faster with better performance, presumably because GoGoD contains professional games, while KGS games are from amateurs and hence a bit noisy. Win rates among the three bots (Table~\ref{tab:win-rate-darkforest}) are consistent with their performances against open source engines.   

\begin{table}[h]
\small
\centering
\begin{tabular}{c||c|c|c|c|c}
             & GnuGo (level 10) & Pachi 10k & Pachi 100k & Fuego 10k & Fuego 100k  \\
\hline\hline
\cite{edinburgh} & 91.0 & - & - & \multicolumn{2}{c}{14.0} \\
\hline
\cite{deepmind}   & 97.2 & 47.4  & 11.0 & 23.3 & 12.5 \\
\hline\hline
\textbf{darkforest} & $98.0 \pm 1.0$ & $71.5 \pm 2.1$ & $27.3 \pm 3.0$ & $84.5 \pm 1.5$ & $56.7 \pm 2.5$ \\
\hline
\textbf{darkfores1} & $99.7 \pm 0.3$  & $88.7 \pm 2.1$ & $59.0 \pm 3.3$ & $93.2 \pm 1.5$ & $78.0 \pm 1.7$ \\
\hline
\textbf{darkfores2}& $\mathbf{100} \pm \mathbf{0.0}$ & $\mathbf{94.3} \pm \mathbf{1.7}$ & $\mathbf{72.6} \pm \mathbf{1.9}$ & $\mathbf{98.5} \pm \mathbf{0.1}$ & $\mathbf{89.7} \pm \mathbf{2.1}$\\  
\end{tabular}
\caption{Win rate comparison against open source engines between our model and previous works. For each setting, 3 groups of 100 games are played. We report the average win rate and standard deviation computed from group averages. All the game experiments mentioned in this paper use komi 7.5 and Chinese rules. Pondering (keep searching when the opponent is thinking) in Pachi and Fuego are on. Note that in~\cite{edinburgh}, they control the time per move as 10 sec/move on 2x 1.6 GHz cores, instead of fixing the rollout number.\protect\footnotemark}
\label{tab:wr-best}
\end{table}
\footnotetext{We also test \emph{darkfores2} against Fuego under this setting, and its win rate is $93\% \pm 1\%$.}

We also compare \emph{darkforest} with a public DCNN model\footnote{From \url{http://physik.de/net.tgz} by Detlef Schmicker. He released the model in Computer-Go forum. See \url{http://computer-go.org/pipermail/computer-go/2015-April/007573.html}}. To create diverse games, moves are sampled according to DCNN softmax probability. We played two sets of 100 games with $100\%$ and $99\%$ win rate. \emph{Darkforest} always wins if sampling from top-1/top-5 moves. 

\begin{table}
\centering
\begin{tabular}{c||c|c|c||c|c|c}
                   & \multicolumn{3}{c}{Move sampled from Top-5} & \multicolumn{3}{c}{Move sampled from Top-300} \\   
\hline
                   &  \emph{darkforest} & \emph{darkfores1} & \emph{darkfores2} & \emph{darkforest} & \emph{darkfores1} & \emph{darkfores2}\\ 
\hline\hline
\emph{darkforest} & $49\%$ & $27\%$ & $17\%$ & $49\%$ & $25\%$ & $15\%$\\
\hline
\emph{darkfores1}  & $70\%$ & $47\%$ & $36\%$ & $69\%$ & $48\%$ & $34\%$\\
\hline
\emph{darkfores2}  & $85\%$ & $55\%$ & $48\%$ & $80\%$ & $59\%$ & $47\%$\\
\end{tabular}
\caption{Win rate among three bots. Each pair plays 100 games. Rows play black. Moves drawn from the DCNN softmax probability.}
\label{tab:win-rate-darkforest}
\end{table}

\textbf{Performance against humans.} We put our bots onto KGS Go server and check their performance against humans over five months period. \emph{Darkforest} became publicly available on Aug 31, 2015. Since then it has played about $2000$ games. Recently we also release the improved version \emph{darkfores1} on Nov 2, 2015, and \emph{darkfores2} after ICLR deadline. All bots become ranked since late November 2015. To score the endgame board situations, we randomly run 1000 trials of default policy to find the dead stones, followed by standard Tromp-Taylor scoring. If all 1000 trials show losing by $10+$ points, they resign.

\begin{figure*}[t!]
    \centering
    \begin{subfigure}
       \centering
       \includegraphics[width=0.48\textwidth]{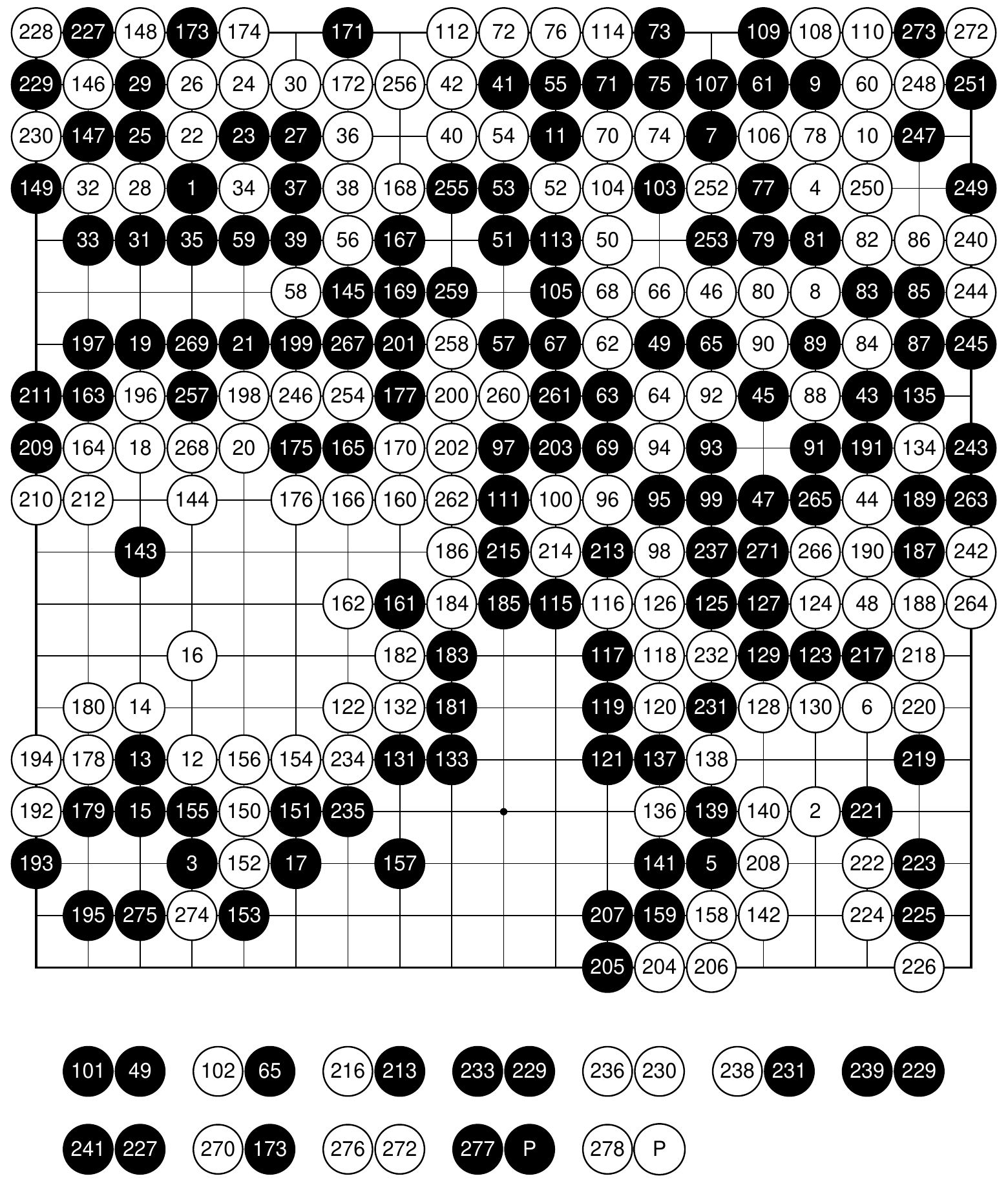}
    \end{subfigure}%
    ~ 
    \begin{subfigure}
        \centering
        \includegraphics[width=0.48\textwidth]{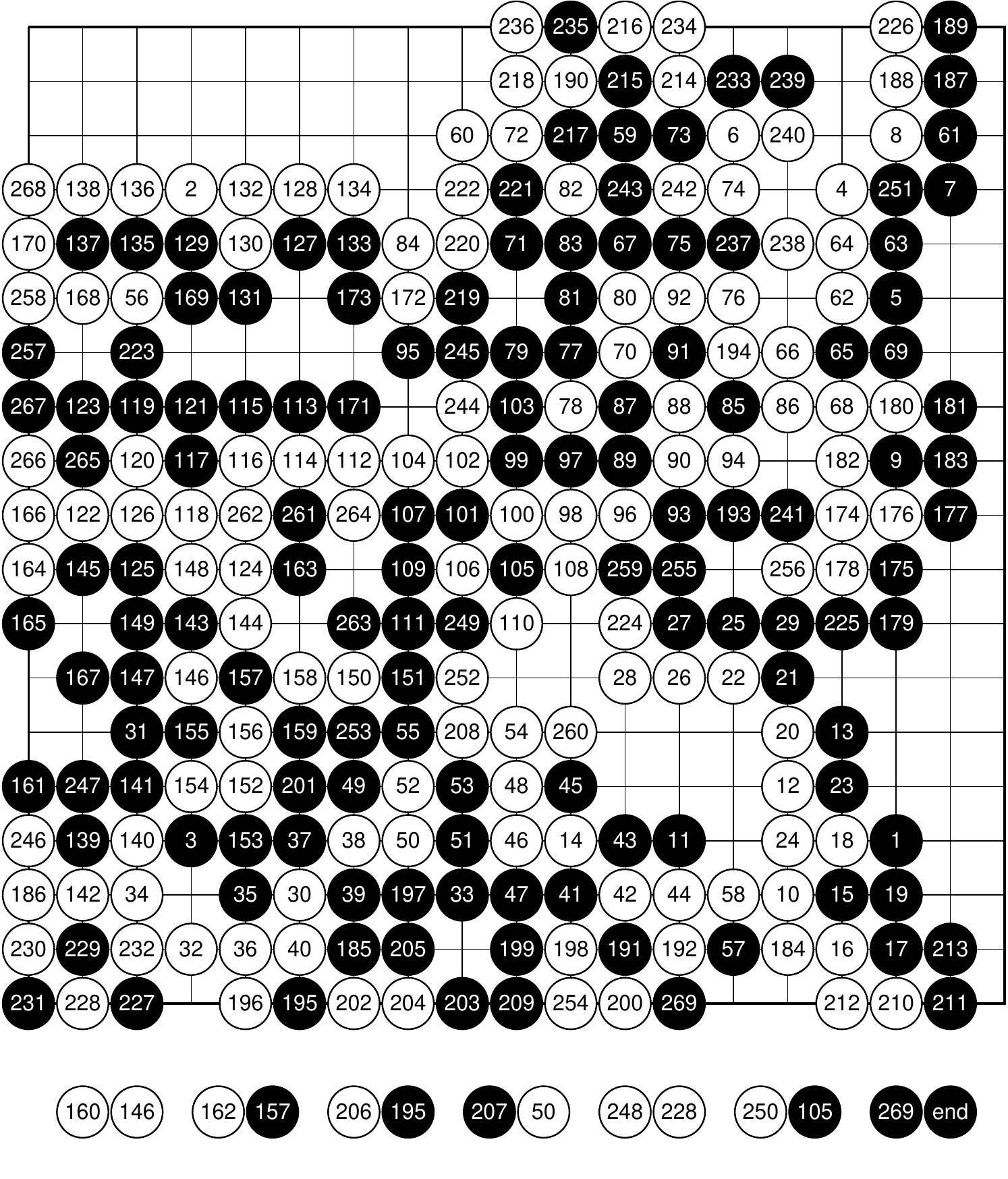}
    \end{subfigure}
    \caption{\textbf{(Left)}: Example game between \emph{darkfores1} (white) and a KGS 1d player \emph{gugun} (black). Move 274 shows the understanding of Ko fight in \emph{darkfores1}. Black has to react in the lower left corner to reinforce the group. In trade, white wins the Ko fight on the top right corner. The result of this game is white win by 7.5 (komi 7.5, Chinese rule). \textbf{(Right)}: Example game between \emph{darkforest} (white) and \emph{darkforest}+MCTS (1000 rollout). \emph{Darkforest} resigned. For concise illustration, we truncated the game when the estimated win rate by MCTS exceeds $0.9$.}
    \label{fig:game-example}
\end{figure*}

All the three pure DCNN bots are quite popular on KGS Go server, playing around $100$ games a day. Once ranked, \emph{darkforest} achieves 1k-1d and \emph{darkfores1} is on strong 2d level, showing the strength of next 3 predictions, consistent with the estimations using free games played in KGS (See Table~\ref{tab:stats-kgs}). The fine-tuned version, \emph{darkfores2}, is on stable 3d level, a very impressive result as pure DCNN models. It even beats a 6d 3 games in a row. We notice that once we open handicap games, their win rates become higher. This is a major improvement upon DCNN developed in~\cite{edinburgh} that holds 4k-5k level, estimated by playing against Go engines. Fig.~\ref{fig:game-example} shows one example game between \emph{darkfores1} and a KGS 1d human player. 

Overall, our bots have a very good understanding of global board situations, and tend to play ``good shapes'' but occasionally fail under local basic life/death situations, e.g., not making a large connected group alive during capturing race, or failing to make two eyes. Rarely they lost in \emph{ladder capture}, a special case in Go (Fig.~\ref{fig:special-situation}(c)) in which one keeps chasing the opponent's stones until the board's border and kill them. Apparently the network failed to capture the ladder pattern due to its rarity in actual games. To handle this, a separate simple search module is added. 

\begin{table}[h]
\small
\centering
\begin{tabular}{c|c||c|c|c|c|c|c|c||c}
        &              &  $<$10k  & 10k - 6k & 5k - 2k & 1k & 1d & 2d & 3d & unranked\\
\hline\hline
\multirow{2}{*}{\textbf{darkforest}} & win rate & $100\%$ & 98.3\% & 90.5\% & 70.4\% & 55.8\% & 50.0\% & 47.1\% & 86.4\% \\
& win/total &  78/78 & 473/481 & 620/685 & 57/81 & 63/113 & 24/48 & 32/68 & 561/649 \\  
\hline
\multirow{2}{*}{\textbf{darkfores1}} & win rate & $100\%$ & 99.1\% & 97.2\% & 87.0\% & 82.5\% & 69.0\% & 62.1\% & 91.1\% \\
& win/total & 17/17 & 218/220 & 345/355 & 60/69 & 47/57 & 20/29 & 18/29 & 357/392 \\
\end{tabular}
\caption{Performance breakdown against different level of players on KGS Go server.}
\label{tab:stats-kgs}
\end{table}

\subsection{Combination with Monte Carlo Tree Search (MCTS)}
We build a standard MCTS framework and study the performance of DCNN+MCTS. \textbf{Tree policy:} Moves are first sorted by DCNN confidences, and then picked in order until the accumulated probability exceeds $0.8$, or the maximum number of top moves are reached. Then we use UCT~\newcite{browne2012survey} to select moves for tree expansion. Note that DCNN confidences are not used in UCT. Noise uniformly distributed in $[0, \sigma]$ is added to the win rate to enforce that search threads quickly diverge and not locked into the same node waiting for DCNN evaluation. This speeds up the tree search tremendously ($\sigma=0.05$ thoroughout the experiments). \textbf{Default policy:} Following Pachi's implementation~\newcite{pachi}, we use 3x3 patterns, opponent \emph{atari} points, detection of \emph{nakade} points and avoidance of self-\emph{atari} for default policy. Note that Pachi's complete default policy yields slightly better performance. Due to the non-deterministic nature of multi-threading, the game between DCNN+MCTS and DCNN is always different for each trial. 

\textbf{Versus Pure DCNN.} In Table~\ref{tab:win-rate-mcts}, \emph{Darkforest}+MCTS gives the highest performance boost over \emph{darkforest}, while boost on \emph{darkfores1} and \emph{darkfores2} is smaller\footnote{We fixed a few bugs from verson 1 of the paper and the baseline win rate versus Pachi/Fuego increases.}. This indicates that MCTS mitigates the weakness of DCNN, in particular for the weaker engine. Another interesting observation is that performance becomes higher if we consider \emph{fewer} top moves in the search. This shows that (1) the top moves from DCNN are really high quality moves and (2) MCTS works better if it digs deep into promising paths. Interestingly, while MCTS with top-2 gives even better performance against pure DCNN, its performance is worse on our KGS version. Finally, setting the minimal number of choices to be more than 1, hurts the performance tremendously. 

\begin{table}[h]
\small
\vspace{-0.2in}
\centering
\begin{tabular}{c||c|c|c}
                            &  darkforest+MCTS & darkfores1+MCTS & darkfores2+MCTS \\
\hline\hline
Vs pure DCNN (1000rl/top-20) & 84.8\%    & 74.0\%    &  62.8\%        \\
Vs pure DCNN (1000rl/top-5)  & 89.6\%    & 76.4\%    &  68.4\%        \\ 
Vs pure DCNN (1000rl/top-3)  & 91.6\%    & 89.6\%    &  79.2\%        \\ 
Vs pure DCNN (5000rl/top-5)  & 96.8\%    & 94.3\%     & 82.3\%        \\
\hline\hline
Vs Pachi 10k (pure DCNN baseline) & 71.5\%  &  88.7\%    &  94.3\% \\ 
\hline\hline
Vs Pachi 10k (1000rl/top-20)   & 91.2\% (+19.7\%)  & 92.0\% (+3.3\%) &  95.2\% (+0.9\%)      \\
Vs Pachi 10k (1000rl/top-5)    & 88.4\% (+16.9\%)  & 94.4\% (+5.7\%) &  97.6\% (+3.3\%)      \\
Vs Pachi 10k (1000rl/top-3)    & 95.2\% (+23.7\%)  & 98.4\% (+9.7\%) &  99.2\% (+4.9\%)      \\
\hline
\hline
Vs Pachi 10k (5000/top-5)      & 98.4\% & 99.6\% & 100.0\% \\
\end{tabular}
\caption{Win rate of DCNN+MCTS against pure DCNN (200 games) and Pachi 10k (250 games).}
\label{tab:win-rate-mcts}
\end{table}

\textbf{Versus Pachi 10k.} With only 1000 rollouts, the win rate improvement over pure DCNN model is huge, in particular for weak models. 5000 rollouts make the performance even better. In particular, darkforest2+MCTS overwhelms Pachi 10k with $100\%$ win rate. Note that for these experiments, Pachi's pondering is turned off (not search during the opponent round).

\textbf{Comparison with previous works.} In comparison, an \emph{asynchronized} version is used in~\cite{deepmind} that achieves $86.7\%$ with 100k rollouts, with faster CPU and GPU (Intel Xeon E5-2643 v2 at 3.50GHz and GeForce GTX Titan Black). The two numbers are not directly comparable since \textbf{(1)} in asynchronized implementations, the number of game states sent to DCNN for evaluation is unknown during 100k rollouts, and \textbf{(2)} Table~\ref{tab:win-rate-mcts} shows that stronger DCNN model benefits less when combined with MCTS. Section~\ref{sec:mcts-intro} gives a detailed comparison between the two implementations. 

\textbf{Evaluation on KGS Go server.} The distributed version, named \emph{darkfmcts3} in KGS Go Server, use \emph{darkfores2} as the underlying DCNN model, runs $75,000$ rollouts on 2048 threads and produces a move every ~13 seconds with one Intel Xeon E5-2680 v2 at 2.80GHz and 44 NVidia K40m GPUs. It uses top-3 predictions in the first 140 moves and switched to top-5 afterwards so that MCTS could have more choices. Pondering is used. Dynamic komi is used only for high handicap games ($\ge$H5). \emph{darkfmcts3} now holds a stable KGS 5d level, on par with the top Go AIs except for AlphaGo~\newcite{nature2016deepmind}, has beaten Zen19 once and hold 1win/1lose against a Korean 6p professional player with 4 handicaps. A version with 110k rollouts and 64 GPUs has won the 3rd place in January KGS Computer Go tournament, where Zen and DolBaram took the 1st and 2nd place\footnote{\emph{darkfmcts3} lost a won game to Zen due to a time management bug, otherwise it would have won 1st place.}, and Abacus~\newcite{graf2015adaptive} took the 4th. With $5000$ rollouts the bot can be run on a single machine with 4 GPUs with 8.8s per move.

\textbf{Weakness.} Despite the involvement of MCTS, a few weakness remains. \textbf{(1)} The top-3/5 moves of DCNN might not contain a critical local move to save/kill the local self/enemy group so local tactics remain weak. Sometimes the bot plays tenuki (``move elsewhere'') pointlessly when a tight local battle is needed. \textbf{(2)} DCNN tends to give high confidences for ko moves even they are useless. This enables DCNN to play single ko fights decently, by following the pattern of playing the ko, playing ko threats and playing the ko again. But it also gets confused in the presence of double ko. \textbf{(3)} When the bot is losing, it plays bad moves like other MCTS bots and loses more. 

\section{Conclusion and Future Work}
In this paper, we have substantially improved the performance of DCNN-based Go AI, extensively evaluated it against both open source engines and strong amateur human players, and shown its potentials if combined with Monte-Carlo Tree Search (MCTS). 

Ideally, we want to construct a system that combines both pattern matching and search, and can be trained jointly in an online fashion. Pattern matching with DCNN is good at global board reading, but might fail to capture special local situations. On the other hand, search is excellent in modeling arbitrary situations, by building a local non-parametric model for the current state, only when the computation cost is affordable. One paradigm is to update DCNN weights (i.e., Policy Gradient~\newcite{sutton1999policy}) after MCTS completes and chooses a different best move than DCNN's proposal. To increase the signal bandwidth, we could also update weights using all the board situations along the trajectory of the best move. Alternatively, we could update the weights when MCTS is running. Actor-Critics algorithms~\newcite{konda1999actor} can also be used to train two models simultaneously, one to predict the next move (actor) and the other to evaluate the current board situation (critic). Finally, local tactics training (e.g., Life/Death practice) focuses on local board situation with fewer variations, which DCNN approaches should benefit from like human players. 

\vspace{0.1in}

\textbf{Acknowledgement} We thank Tudor Bosman for building distributed systems, Rob Fergus and Keith Adams for constructive suggestions, and Vincent Cheung for engineering help.

\bibliography{reference}
\bibliographystyle{iclr2016_conference}

\end{document}